\documentclass[letterpaper,journal]{IEEEtran}
\usepackage{amsmath,amsfonts}
\usepackage{algorithmic}
\usepackage{algorithm}
\usepackage{array}
\usepackage[caption=false,font=normalsize,labelfont=sf,textfont=sf]{subfig}
\usepackage{textcomp}
\usepackage{stfloats}
\usepackage{url}
\usepackage{verbatim}
\usepackage{graphicx}
\usepackage{cite}
\usepackage{subfig}
\usepackage{nccmath, graphicx}
\usepackage{multirow}
\usepackage{makecell}
\usepackage{adjustbox}
\usepackage{hhline}
\usepackage{balance}

\begin{document}

\title{Unified Contrastive Fusion Transformer for Multimodal Human Action Recognition }
\author{Kyoung Ok Yang, Junho Koh and Jun Won Choi$^*$
\thanks{* Corresponding author: Jun Won Choi.}
\thanks{K. Yang is with Department of Artificial Intelligence, Hanyang University, 222, Wangsimni-ro, Seongdong-gu, Seoul, 04763, Korea (email: koyang@spa.hanyang.ac.kr)}%
\thanks{J. Koh is with Department of Electrical Engineering, Hanyang University, 222, Wangsimni-ro, Seongdong-gu, Seoul, 04763, Korea (email: jhkoh@spa.hanyang.ac.kr)}%
\thanks{J. W. Choi is with Department of Electrical Engineering and Graduate School of Artificial Intelligence, Hanyang University, 222, Wangsimni-ro, Seongdong-gu, Seoul, 04763, Korea (email: junwchoi@hanyang.ac.kr)}
}

\markboth{}%
{Unified Contrastive Fusion Transformer for Multimodal Human Action Recognition}


\maketitle

\begin{abstract}
Various types of sensors have been considered to develop human action recognition (HAR) models. Robust HAR performance can be achieved by fusing multimodal data acquired by different sensors. In this paper, we introduce a new multimodal fusion architecture, referred to as Unified Contrastive Fusion Transformer (UCFFormer) designed to integrate data with diverse distributions to enhance HAR performance. Based on the embedding features extracted from each modality, UCFFormer employs the Unified Transformer to capture the inter-dependency among embeddings in both time and modality domains. We present the Factorized Time-Modality Attention to perform self-attention efficiently for the Unified Transformer. UCFFormer also incorporates contrastive learning to reduce the discrepancy in feature distributions across various modalities, thus generating semantically aligned features for information fusion. Performance evaluation conducted on two popular datasets, UTD-MHAD and NTU RGB+D, demonstrates that UCFFormer achieves state-of-the-art performance, outperforming competing methods by considerable margins. 
\end{abstract}


\section{Introduction}\label{sec1}
\IEEEPARstart{H}{uman} Action Recognition (HAR) is a process that involves the automatic identification and classification of human actions based on sensor data. 
HAR has a wide range of applications, including healthcare monitoring \cite{liu2022overview}, fitness tracking \cite{pathan2019machine}, action analysis, gesture-based interfaces \cite{saini2020human}, and context-aware systems \cite{fan2022context}. The ability to automatically recognize and classify human activities based on sensor measurements has the potential to enhance various domains and improve user experiences. 

Various sensors such as video camera, wearable devices \cite{kumari2017increasing},  and environmental sensors \cite{cippitelli2017human} can be utilized to acquire data for HAR. 
Based on the sensor types, HAR  methods can be broadly classified into two categories: those based on visual data and those based on non-visual data. Visual data is acquired from camera sensors, including video, depth, and infrared cameras.
These camera sensors capture high-resolution images that depict the movement and posture of individuals. Deep learning models, especially convolutional neural networks (CNNs), have been widely employed for HAR using visual data \cite{taylor2010convolutional, qiu2017learning, carreira2017quo, feichtenhofer2019slowfast}.
On the other hand, non-visual data can be acquired from various sensors including accelerometers, gyroscopes, magnetometers, and pressure sensors. These sensors capture non-visual data related to the body's physical movements. Devices like smartwatches \cite{diete2019vision}, fitness trackers \cite{chen2014home}, and smartphones \cite{khandnor2017survey} integrate such sensors, allowing users to continually monitor their activities. Various deep learning architectures for modeling sequential data  have been used to conduct HAR based on non-visual data \cite{zhao2018deep,al2018hierarchical,gao2021danhar}.

The use of multimodal data is motivated by several factors. First, different sensor modalities capture different aspects of human motion and provide complementary information that enhances the accuracy of action recognition \cite{dang2020sensor}. Second, using multiple sensor modalities can offer redundancy, which can improve the robustness of action recognition \cite{ravi2016deep}. Third, multimodal data can provide a more comprehensive representation of human action, improving generalization and adaptability of HAR models \cite{ravi2016deep, atrey2010multimodal}. By capturing different aspects of human motion across various sensor modalities, the models can be trained to generalize across diverse scenarios, users, and environments.

However, utilizing multimodal data in HAR poses some challenges. One of the main challenges is establishing a joint representation for multimodal data that captures complex relational semantics. To maximize the benefits of sensor fusion, HAR models should produce the complementary yet semantically well aligned features from each modality.
Another challenge is optimally  combining the data from different sensors and modalities. The varied forms of data are typically noisy, intricate, and exhibit high variability. These challenges necessitate an efficient sensor fusion algorithm capable of modeling their relational structure and integrating data adaptively based on its quality.

To date, various multimodal fusion architectures have been proposed  for HAR. 
In \cite{moencks2019adaptive}, Meoncks et al.  proposed adaptive feature processing for HAR using RGBD and LiDAR, and in  \cite{shahroudy2017deep}, Shahroudy et al.  presented the shared specific feature factorization network to combine multimodal signals. Hierarchical Multimodal Self-Attention (HAMLET) \cite{islam2020hamlet} combined global and local features extracted from RGB and IMU through Transformer.
Multitask Learning-based Guided Multimodal (MuMu) \cite{islam2022mumu} integrated a multi-task learning strategy to a multimodal fusion network. 
Vision-to-Sensor Knowledge Distillation (VSKD) strategy \cite{ni2022cross} proposed the network for transferring the knowledge from video modality to inertial modality.
MAVEN \cite{islam2022maven} enhanced the performance of multimodal fusion by using a memory-augmented recurrent network and aligning representations using an attention mechanism.

In this study, we introduce  a new  unified multimodal fusion framework  for HAR, referred to as the Unified Contrastive Fusion Transformer (UCFFormer). This framework effectively combines multimodal data of varying distributions, including both visual and nonvisual data. The UCFFormer is different from the existing fusion architectures in these two aspects. First, UCFFormer derives joint representation of multimodal sensor data using a unified Transformer architecture, which has been successfully used for vision-language modeling  lately \cite{zhou2020unified}. It first extracts the embedding features of the same size from each multimodal input and encodes them via Multimodal Self-attention capturing pairwise interactions across both time and modality domains. 
We introduce an efficient implementation of Multimodal Self-attention for the unified Transformer. We devise a Factorized Time-Modality Self-attention to independently encode the embeddings in both time and modality domains. Two strategies are proposed for  factorized attention: parallel and sequential factorization. While parallel factorization conducts self-attention across both time and modality domains simultaneously, sequential factorization alternates between them in each Transformer layer.

Next, we propose a feature alignment strategy based on a contrastive learning approach. By minimizing the cosine similarity metric defined between the embedding vectors of different modalities, the proposed method can mitigate the domain gap between the modalities and thereby boost the effectiveness of multimodal feature fusion. As a result, the proposed Multimodal Contrastive Alignment Network (MCANet) generates embedding features that are more coherent and semantically aligned. 

We evaluate the performance of UCFFormer on two widely used multimodal HAR datasets, UTD-MHAD \cite{chen2015utd} and NTU RGB+D \cite{shahroudy2016ntu}. Our evaluation demonstrates that UCFFormer outperforms existing HAR models by significant margins, recording a new state-of-the-art performance. In particular, UCFFormer achieves remarkable 99.99\% Top-1 accuracy on UTD-MHAD dataset. 

The key contributions of this study are summarized as follows:
\begin{itemize}
    \item We present a novel multimodal fusion network called UCFFormer. Our approach leverages a unified Transformer to enhance the embedding features extracted from different sensor data modalities. 
    This structure enables sensor fusion whose design is agnostic to the type and number of multimodal inputs, without the need for specific custom designs for different modality types. 
    \item We introduce the Factorized Time-Modality Self-attention for the efficient encoding of embedding features. To this goal, we present two distinct factorization strategies. 
    \item We further enhance the effects of sensor fusion by aligning the embedding features semantically using contrastive learning. To the best of our knowledge, we are the first to demonstrate the efficacy of contrastive learning in multimodal HAR. Our contrastive learning-centric alignment technique can be seamlessly integrated into any feature fusion module.
    \item Our UCFFormer achieves the state-of-the-art performances on UTD-MHAD \cite{chen2015utd} and NTU RGB+D \cite{shahroudy2016ntu} datasets. 
\end{itemize}

\section{Related Works}\label{sec2}
\subsection{Multimodal Fusion Methods for HAR}
Early studies in multimodal fusion presented a late fusion approach which combined classification results obtained from each modality to obtain a final prediction. 
In \cite{dawar2018action,dawar2018data}, Dawar et al. encoded RGB images using CNN and inertial sensor data using Long Short Term Memory (LSTM) and combined the resulting classification outcomes using the weights derived from the decision fusion scores. In \cite{imran2020evaluating}, 1D CNN, 2D CNN, and Recurrent Neural Network (RNN) were employed to predict action class based on Gyroscope data, RGB data, and human joint pose data, respectively and the classification results were combined. 
In \cite{zou2019wifi}, WiVi utilized a CNN backbone to represent WiFi signal and a C3D backbone \cite{tran2015learning} to process RGB data. These were then integrated at the decision level through an ensemble fusion model.

Several studies have explored achieving information fusion at the intermediate feature level.  
In \cite{long2018multimodal}, a Keyless Attention method was presented to aggregate the features extracted from multimodal data.
HAMLET \cite{islam2020hamlet} employed Hierarchical Multimodal Self-attention to obtain action-related spatio-temporal features. MuMu \cite{islam2022mumu} was trained to conduct multiple tasks, i.e.,  the target task of HAR and an auxiliary task of human action grouping. The auxiliary task assists the target task in extracting appropriate multimodal representations.
VSKD \cite{ni2022cross} utilized ResNet18 as a teacher network for video data and employed multi-scale TRN \cite{zhou2018temporal} with BN-Inception as a student network for inertial data.  Distance and Angle-wise Semantic Knowledge (DASK) loss was proposed to account for the modality differences between the vision and sensor domains.
MAVEN \cite{islam2022maven}  employed the feature encoders to generate modality-specific spatial features that were subsequently stored in memory banks. The memory banks were used to capture long-term spatiotemporal feature relationships. 

Our UCFFormer is different from aforementioned methods in that a joint representation of multimodal data is found through the time-modality factorized self-attention of the unified Transformer and using contrastive learning framework.

\subsection{Contrastive Learning}
Contrastive learning is a type of the self-supervised learning tasks that provides a means of understanding the differences between representations. This stems from the work of Bromley et al. who introduced the concept of a Siamese Network, consisting of two identical networks that share weights for metric learning \cite{bromley1993signature}. Specifically, contrastive learning examines which pairs of data points are similar and different to learn high-level data features before performing classification or segmentation tasks.

Early studies used a contrastive learning framework to learn invariant mappings for recognition using contrastive pair loss in discrimination models \cite{chopra2005learning,hadsell2006dimensionality}. Inspired by triplet loss, recent studies applied feature extraction methods that minimize the distance between representations of similar positive pairs and maximize the distance between representations of different negative pairs \cite{weinberger2009distance,collobert2008unified,chechik2010large}.

Recently, contrastive learning has widely been used for various image classification tasks.
Momentum Contrast (MoCo) \cite{he2020momentum,chen2020improved} used a momentum-updated encoder to produce representations of negative samples, providing a large and consistently maintained set of negatives for contrastive loss calculations.
SimCLR \cite{chen2020simple,chen2020big} learned representations by maximizing the similarity between different augmentations of the same data sample while minimizing the similarity between different samples.
Bootstrap Your Own Latent (BYOL) \cite{grill2020bootstrap} achieved self-supervised image representation learning without using negative samples by creating two augmented views of the same image.

\subsection{Unified Transformer for Multimodal Task}
Transformer architectures have achieved significant success in machine learning tasks, including natural language processing \cite{kenton2019bert,brown2020language} and computer vision \cite{dosovitskiy2020image}. However, they have mainly been limited to tasks within a single domain or specific multimodal domains. To overcome this challenge, a Unified Transformer model \cite{hu2021unit} has been proposed as a foundation model for multimodal data. 
The Unified Transformer consists of transform encoders jointly encoding multimodal inputs, and a transform decoder over the encoded input modalities, followed by task-specific outputs applied to the decoder hidden states to make final predictions for each task. The Unified Transformer handles multi-tasking and multimodal in a single model with fewer parameters, moving toward general intelligence. 

The Unified Transformer has been used for a variety of tasks that involve multimodal data.
UFO \cite{wang2021ufo} used a single transformer network for multimodal data and implemented multi-task learning during vision-language pre-training. The same transformer network was used as an image encoder, as a text encoder, or as a fusion network in the different pre-training tasks.
UniFormer \cite{li2021uniformer} unified 3D convolution and spatio-temporal self-attention to generate a transformer embedding vector. UniFormer's relation aggregator handles both spatio-temporal redundancy and dependency by learning local and global token correlations in shallow and deep layers, respectively.
UniTranSeR \cite{ma2022unitranser} embedded the multimodal features into a unified Transformer semantic space to prompt inter-modal interactions, and then employed a Feature Alignment and Intention Reasoning layer to perform cross-modal entity alignment.

\section{Unified Contrastive Fusion Transformer (UCF-Former)}\label{sec3}
\begin{figure*}[tbh]
    \centering
    \includegraphics[width=0.99\textwidth]{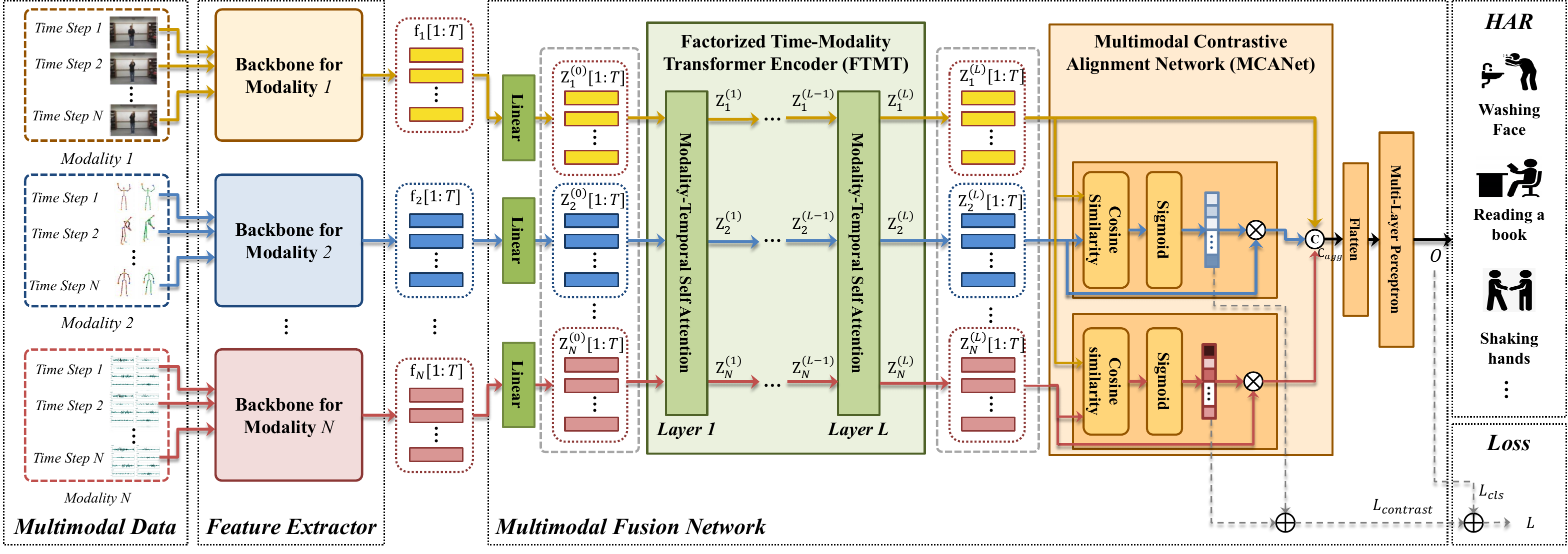}
    \caption{\textbf{Overall Architecture} : The proposed UCFFormer first represents each raw multimodal data in a shared feature space. FTMT encodes these embedding features, capturing their dependencies within the time-modality domain using the Unified Transformer. Subsequently, MCANet refines these embeddings by aligning them across modalities through contrastive learning. These enhanced embeddings are then aggregated, leading to the final classification results.}
    \label{fig1:Overall_Architecture}
\end{figure*}

\subsection{Overview}
Figure \ref{fig1:Overall_Architecture} depicts the overall structure of the proposed UCFFormer. The proposed UCFFormer fuses the information from $N$ multimodal sensors. First, $N$ separate backbone networks are employed to extract sequential feature vectors of length $T$ from each modality. The feature vectors are linearly projected to produce the $NT$ embedding vectors of the same size. The projected embedding vectors serve as basic semantic elements represented in time and modality domain.
Factorized Time-Modality Transformer (FTMT) encodes the embedding vectors using a unified Transformer architecture. The unified Transformer models both intra-modality and inter-modality interactions simultaneously to produce the updated embedding vectors. To facilitate effective interaction modeling, FTMT employs a {\it factorized self-attention mechanism} that conducts the encoding process separately in temporal and modality domains.

Next, the Multimodal Contrastive Alignment Network (MCANet) combines the features generated by FTMT.
Among $N$ multimodal sensors, we designate one as main modality sensor and the others as $N-1$ sub-modality sensors. MCANet boosts the effect of feature fusion by aligning the sub-modality features with those of the main modality through contrastive learning. Finally, the combined features are passed through a multi-layer perceptron (MLP) layer followed by a softmax layer to generate the final classification result.

\subsection{Setup and notations}
We make the assumption that $N$ sensor data are temporally synchronized. To achieve this, we resample the data such that their sampling rates are all identical. Our model takes $T$ consecutive samples of all modality data in a sequential manner. We denote $T$ signal samples acquired from $N$ modality sensors as $x_{1}[1:T], x_{2}[1:T],..., x_{N}[1:T]$, where $x_{n}[1:T]=\{x_{n}[1],..., x_{n}[T]\}$. The dimension of each sample is different for each modality. 

\subsection{Multimodal Feature Extractor}
Different backbone networks are employed to extract feature vectors from each modality data. Suppose that the feature vectors $f_{n}[1:T]$ are obtained from the input data $x_{n}[1:T]$.
Then, we apply linear projection to map the feature embedding vectors into the embedding vectors of common size $d$, i.e.,
\begin{equation}
    \begin{split}
     Z_{1}[1:T] &= W_{1}\cdot f_{1}[1:T] \\
   & \vdots \\
    Z_{N}[1:T] &= W_{N}\cdot f_{N}[1:T].
    \end{split}
  \end{equation}
These $NT$ embedding vectors form the basis for finding the joint representation of multimodal data.

\subsection{Factorized Time-Modality Transformer Encoder}
FTMT encoder encodes the embedding vectors capturing their high-level inter-dependencies. 
To this goal, we utilize the unified Transformer, which has been utilized for vision-language multimodal data modeling  \cite{zhou2020unified}.  The unified Transformer leverages Transformer Self-attention to encode the embedding vectors across different modalities and time steps.

However, given $NT$ embedding vectors, FTMT requires the computational complexity of $\mathcal{O}(N^2 T^2)$ and high capacity network for learning all possible pairwise relations among $NT$ elements. To cope with this issue, we employ the Factorized Time-Modality Self-attention, which conducts self-attention across time and modality domains  independently. We present two distinct versions of factorized self-attention, which are differentiated by their respective arrangements of time-domain and modality-domain self-attention operations.

\subsubsection{Module 1: Simultaneous Time-Modality Factorization }
\begin{figure*}[tbh]
    \centering
    \captionsetup[subfloat]{labelfont=footnotesize,textfont=footnotesize}
    \subfloat[\textbf{Simultaneous Time-Modality Factorization (FTMT-Sim)}]{\includegraphics[width=0.50\textwidth]{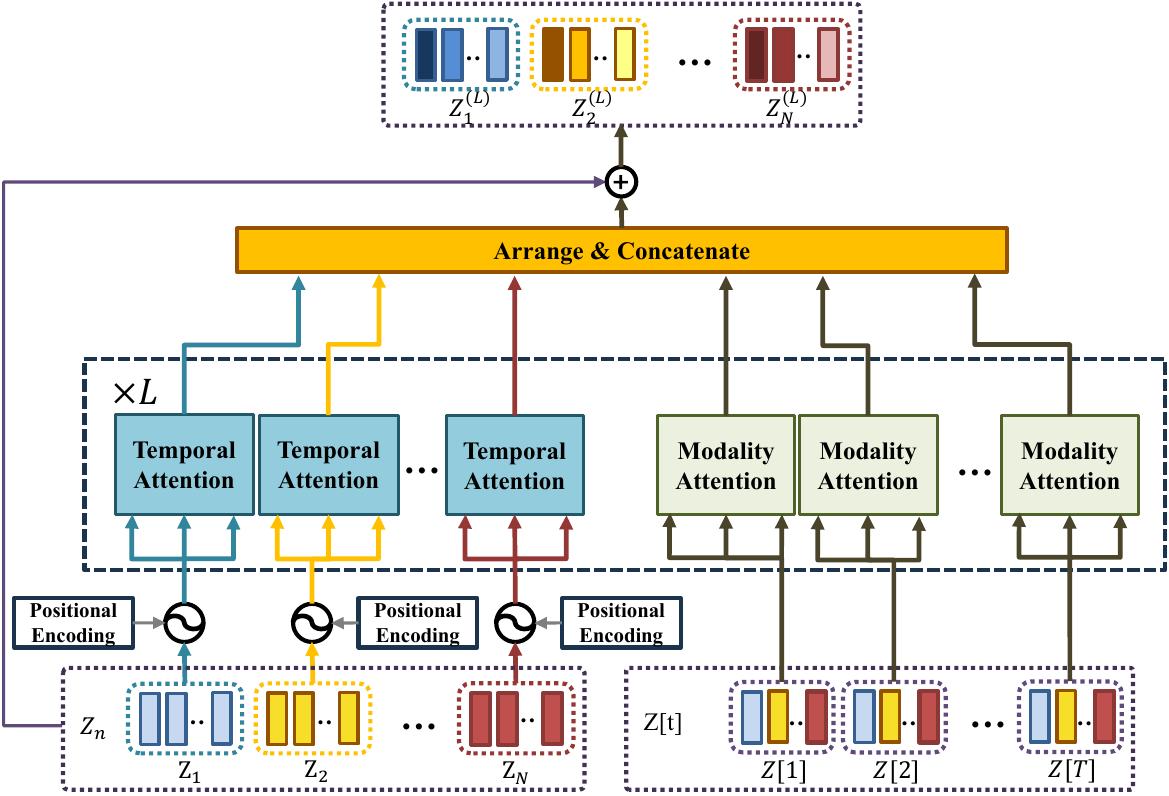}}
    \quad\quad
    \subfloat[\textbf{Sequential Time-Modality Factorization (FTMT-Seq)}]{\includegraphics[width=0.3\textwidth]{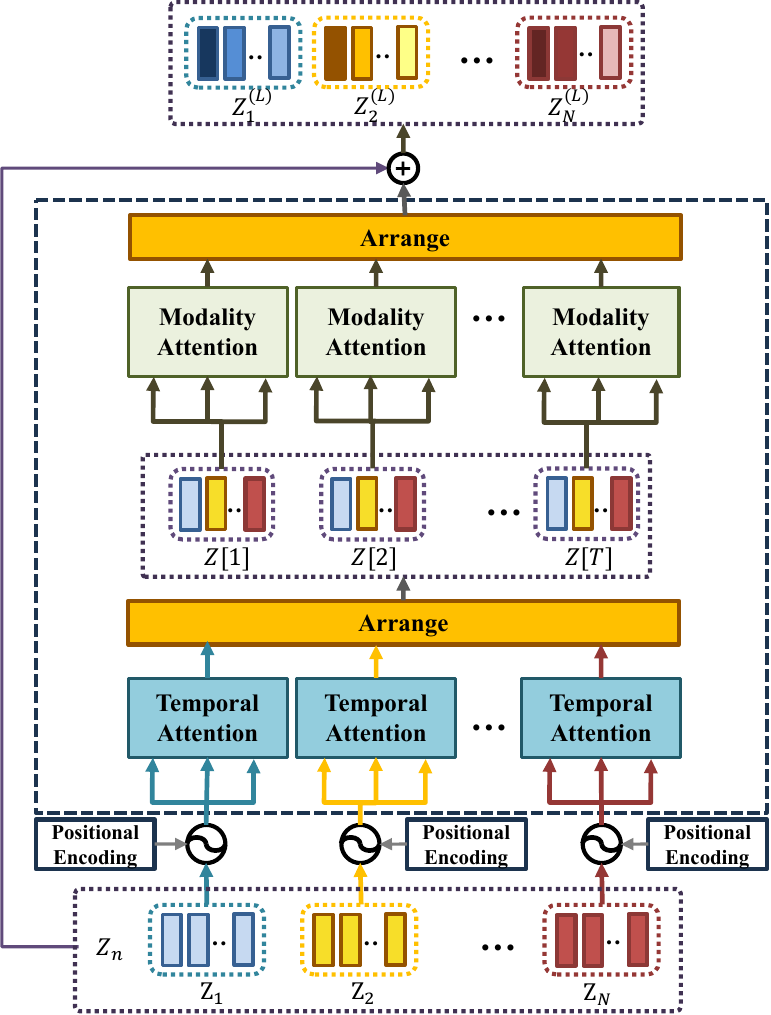}}
    \caption{\textbf{Factorized Time-Modality Self-attention} : (a) FTMT-Sim processes the embedding features in both time and modality domains concurrently, (b) FTMT-Seq alternates the encoding process between the time-domain and the modality-domain in a sequential manner. }
    \label{fig2:FSC}
\end{figure*}

The Simultaneous Time-Modality Factorization (FTMT-Sim) approach conducts self-attention operations concurrently in the time and modality domains and merges the encoded features in the final attention layer. Figure \ref{fig2:FSC} (a) depicts the structure of FTMT-Sim. For each time step $t$, the embedding vectors across different modalities are packed into a matrix $Z[t]=[Z_{1}[t],...,Z_{N}[t]]$. Similarly, the embedding vectors across time steps are packed into a matrix $Z_{n}=[Z_{n}[1],...,Z_{n}[T]]$. Then, self-attention is applied independently to each axis in parallel. First, self attention across modalities is performed as
\begin{equation}
    \begin{split}
    Z^{(l+1)}[t] &= \mathrm{ModalityAttention}(Z^{(l)}[t]) \\ 
    &= \mathrm{Softmax}\left( \frac{Q^{(l)}[t]((K^{(l)}[t])^{T}V^{(l)}[t])}{\sqrt{d_{k}}} \right),
    \end{split}    
\end{equation}
where 
\begin{equation}
    \begin{split}
    Q^{(l)}[t] &= Z^{(l)}[t] \cdot W^{(l)}_{q}  \\
    K^{(l)}[t] &= Z^{(l)}[t] \cdot W^{(l)}_{k}  \\
    V^{(l)}[t] &= Z^{(l)}[t] \cdot W^{(l)}_{v},
    \end{split}
\end{equation}
$l \in [1,L] $ is the index of attention layer and $W^{(l)}_{q} \in \mathbb{R}^{d \times d_{k}}$, $W^{(l)}_{k} \in \mathbb{R}^{d \times d_{k}}$, and $W^{(l)}_{v} \in \mathbb{R}^{d \times d_{v}}$ are the linear weights, where $d_{k}$ and $d_{v}$ represent the dimensions of the keys and values, respectively. For brevity, we exclude the notation for multi-head attention.
In our implementation, both $d_{k}$ and $d_{v}$ are set to 64.

Next, the positioning encoding is applied to the embedding vectors in $Z_{n}$ \cite{vaswani2017attention} for each modality index $n$.  Then, self-attention across time steps is performed as
\begin{equation}
    \begin{split}
    Z^{(l+1)}_{n} &= \mathrm{TemporalAttention}(Z^{(l)}_{n}) \\ 
    &=  \mathrm{Softmax}\left( \frac{Q_n^{(l)}((K_n^{(l)})^{T}V_n^{(l)})}{\sqrt{d_{k}}} \right)\,
    \end{split}    
\end{equation}
where 
\begin{equation}
    \begin{split}
    Q_n^{(l)} &= Z^{(l)}_{n} \cdot V^{(l)}_{q} \\
    K_n^{(l)} &= Z^{(l)}_{n} \cdot V^{(l)}_{k} \\
     V_n^{(l)} &= Z^{(l)}_{n} \cdot V^{(l)}_{v},
     \end{split}
\end{equation}
 $W^{(l)}_{n,q} \in \mathbb{R}^{d \times d_{k}}$, $W^{(l)}_{n,k} \in \mathbb{R}^{d \times d_{k}}$, and $W^{(l)}_{n,v} \in \mathbb{R}^{d \times d_{v}}$ are the linear weights.

After $L$ attention layers, the attention values $Z^{(L)}[1],..., Z^{(L)}[T]$ and $Z^{(L)}_{1},...,Z^{(L)}_{N}$ are arranged, concatenated and linearly projected, resulting in the generation of the final $NT$ features $C_{1}[1:T],...,C_{N}[1:T]$. 
Note that the skip connection, originating from the initial values $Z^{(0)}_{n}$ and $Z^{(0)}[t]$, is incorporated during the process of generating the final features. 

\subsubsection{Module 2: Sequential Time-Modality Factorization (FTMT-Seq)}
Unlike FTMT-Sim, the Sequential Time-Modality Factorization (FTMT-Seq) approach applies the self-attention operations in time axis and modality axis one by one. First, after positional encoding is performed, the time-domain self-attention operation is performed  as
\begin{equation}
    Y^{(l)}_{n} = \mathrm{TemporalAttention}(Z_n^{(l)})
\end{equation}
Then, the output $Y^{(l)}_{1},...,Y^{(l)}_{N}$ are rearranged to $Y^{(l)}[1],...,Y^{(l)}[T]$.
Then, the modality-domain self-attention follows 
\begin{equation}
    Z^{(l+1)}[t] =  \mathrm{ModalityAttention}(Y^{(l)}[t]). 
\end{equation}
After passing through $L$ attention layers and skip connection from the input, FTMT-Seq produces the final features  $C_{1}[1:T],...,C_{N}[1:T]$.

\subsection{Multimodal Constrastive Alignment Network}
Utilizing the output of FTMT, $C_{1}[1:T],...,C_{N}[1:T]$, MCANet employs Weighted Multimodal Feature Fusion that combines the multi-modal features.  In our framework, we let $C_{1}[1:T]$ be the main modality features and the rest be the sub-modality features.
The primary goal of this approach is to selectively aggregate the information that is pertinent to the main modality data from the sub-modality data. 
For each time step $t$, Weighted Multimodal Feature Fusion combines the multimodal features as
\begin{align}
        C_{agg}[t] &= C_{1}[t] + \sum_{n=2}^{N} \mathrm{sigmoid}(\mathrm{sim}(C_{1}[t], C_{n}[t])) C_{n}[t],   \end{align}
where $\mathrm{sim}(A,B)$ denotes the cosine similarity measure
\begin{align}
    \begin{split}
    \mathrm{sim}(A,B) = \frac{A^{T} \cdot B}{\lVert A \rVert_2 \lVert B \rVert_2}. 
    \end{split}
\end{align}
Note that the sub-modality features $C_{n}[t]$ $n \neq 1$ are weighted according to their similarity with the main modality features  $\mathrm{sigmoid}(\mathrm{sim}(C_{1}[t], C_{n}[t]))$. 
Additionally, we employ a contrastive learning framework to reduce feature discrepancy across different modalities. By minimizing a contrastive loss that captures the dis-similarities between the different modalities during training, we can make our weighted feature aggregation more effective. Details on the contrastive loss will be discussed in the following subsection.


Finally, MCANet flattens the matrix $[{C}_{agg}[1], ..., {C}_{agg}[T]] $ into a single-dimensional vector. The flattened vector passes through the MLP followed by softmax layer, generating the final classification scores $O\in \mathbb{R}^{n_{cls}}$, where $n_{cls}$ is the number of action classes. Each element in the vector $O$ represents the probability score assigned to a specific action class. By analyzing these probabilities, the model can determine the most likely class for a given input.

\subsection{Loss Function}
As depicted in Figure \ref{fig1:Overall_Architecture}, the loss function used to train our entire network consists of the contrastive loss term $\mathcal{L}_{contrast}$ and the cross entropy loss term $\mathcal{L}_{cls}$, i.e.,  
\begin{equation}
\mathcal{L} = \mathcal{L}_{cls} + \alpha \mathcal{L}_{contrast},
\end{equation}
where $\alpha$ is the regularization parameter. In our approach, we set $\alpha$ to 0.2.

The contrastive loss term $\mathcal{L}_{contrast}$ aims to encourage the alignment of features among different modalities. It quantifies the dissimilarity or disparity between the main modality and the $N-1$ sub-modalities, i.e.,  
\begin{equation}
\label{eq:diff_loss}
    \mathcal{L}_{contrast} = - \sum_{n=2}^{N} \log \left(\frac{1}{T}\sum_{t=1}^{T}\mathrm{sim}(C_{1}[t],C_{n}[t])\right)
\end{equation}
By using the contrastive loss term, the model can generate semantically consistent representations. 

The cross entropy loss term $\mathcal{L}_{cls}$ is given by
\begin{equation}
\mathcal{L}_{cls} = - \sum_{i = 1}^{n_{cls}}{T_{i}\log(O_{i})},
\end{equation}
where $O_i$ is the $i$th element of $O$ and $T_i$ is the $i$th element of ground truth one hot vector.  

\section{Implementation of Multimodal HAR System}
In this section, we present design variants of UCFFormer, specifically tailored for the sensor configurations provided in two datasets: UTD-MHAD \cite{chen2015utd} and NTU RGB+D \cite{shahroudy2016ntu}.

\subsection{Implementation of UCFFormer on UTD-MHAD dataset}
UCFFormer is designed to fuse three different modalities including RGB video,  skeleton data and inertial sensor signals provided by UTD-MHAD dataset.  RGB video consists of length-$T$ sequence of video frames, i.e.,  $x_{1}[1:T] \in \mathbb{R}^{T \times H \times W}$, where $H$ and $W$ are the height and width of a video frame. The skeleton data  consists of 3D coordinates of $J$ joints, i.e., $x_{2}[1:T] \in \mathbb{R}^{T \times J \times 3}$, where $J$ is the number of joints. The data from the inertial sensors is divided into segments, with each segment being processed individually. The inertial sensor data is also processed in a similarly way, yielding vectors of the dimension $S$,  $x_{3}[1:T] \in \mathbb{R}^{T \times S}$.

We tried two different camera backbones; the ResNet50 model and  the Temporal Shift Module (TSM) \cite{lin2019tsm}. ResNet50 was used to to encode every frame of the RGB video while TSM was used to encode sequential RGB video.  We specifically selected ResNet50 to ensure fair comparison with several latest methods \cite{imran2016human,liu2018rgb,islam2020hamlet,islam2022mumu} that utilized ResNet50 on the UTD-MHAD dataset. We also tried a stronger backbone, TSM at the cost of higher computational complexity.  These backbone networks take video frames $x_1[1:T]$ as input and produces the sequence of feature maps $f_1[1:T]$. We encoded the skeleton data $x_2[1:T]$ using a Spatio-Temporal Graph Convolutional Network (STGCN) \cite{yu2018spatio}. Lastly, a DeepConvLSTM \cite{singh2020deep} was employed to encode the $T$ segments of inertial sensor data $x_{3}[1:T]$ and produce the feature vectors $f_{3}[1:T]$. 
The remaining steps follow the procedure described previously. 
In contrastive learning setup, RGB video acted as a primary modality, while the remaining data sources served as sub-modalities.

\subsection{Implementation of UCFFormer on NTU RGB+D dataset}
We consider two modalities: RGB images and skeleton data on NTU RGB+D dataset. The data preprocessing step was performed similarly as in UTD-MHAD dataset. To encode the RGB image sequence, we employ TSM as a video backbone network. 
The samples of NTU RGB+D dataset capture both individual actions and interactions among multiple individuals. 
From this perspective, we processed skeleton data for each of $P$ persons and constructed the tensor, 
$x_{2}[t] \in \mathbb{R}^{T \times J \times P \times 3}$.
The skeleton data was processed using the same backbone architecture used for the UTD-MHAD dataset. In contrastive learning setup, RGB video was selected as a primary modality.

\section{Empirical Results}\label{sec4}
\newcolumntype{D}{>{\centering\arraybackslash}p{9.5em}}
\newcolumntype{M}{>{\centering\arraybackslash}p{8.0em}}
\newcolumntype{A}{>{\centering\arraybackslash}p{4.5em}}
In this section, we evaluate the performance of the proposed UCFFormer on two widely used datasets: UTD-MHAD \cite{chen2015utd} and NTU RGB+D \cite{shahroudy2016ntu}. 

\subsection{Dataset} 

\textbf{UTD-MHAD:} The UTD-MHAD dataset comprises 27 human actions performed by eight subjects. It offers data in four distinct modalities: RGB video, depth video, skeletal joint positions, and inertial sensor signals.
For evaluating performance, we utilized the Top-1 accuracy metric. In this dataset, we adopted the cross-subject evaluation method widely used to evaluate the performance of HAR models \cite{chen2015utd}. This approach divides subjects into different groups for training versus testing to assess the model's ability to generalize to new, unseen subjects. Specifically, we used a dataset comprising 8 subjects for this experiment. Among them, 6 subjects were utilized for training, 1 subject for validation, and 1 subject for testing.

\textbf{NTU RGB+D:} The NTU RGB+D dataset is an extensive collection of RGB video, depth video and skeleton data, consisting of 56,880 action samples performed by 40 subjects across 60 action classes.  In this dataset, two evaluation methods, namely cross-subject (CS) and cross-view (CV), have been widely employed \cite{shahroudy2016ntu}. For the CS evaluation method, 40 samples were randomly selected as training and testing groups, with the training IDs being 1, 2, 4, 5, 8, 9, 13, 14, 15, 16, 17, 18, 19, 25, 27, 28, 31, 34, 35, and 38. The remaining subjects were reserved for testing, resulting in 40,320 for training and 16,560 samples for testing. On the other hand, for CV evaluation, samples from cameras 2 and 3 were used for training, while samples from camera 1 were used for testing. The resulting sets comprised 37,920  for training and 18,960 samples for testing.

\subsection{Experimental Setup}

\subsubsection{Implementation Details}
We present the implementation details of UCFFormer. 
The performance of UCFFormer was evaluated across four different configurations. Based on the two types of proposed  factorized attention, UCFFormer is divided into UCFFormer-Sim and UCFFormer-Seq. Additionally, when using the Image Backbone ResNet50, it was denoted as R, and when using the Video Backbone TSM, it was denoted as T. Therefore, the four tested types were denoted as UCFFormer-Sim-R, UCFFormer-Seq-R, UCFFormer-Sim-T, and UCFFormer-Seq-T.

\textbf{UTD-MHAD:} 
In the case of UTD-MHAD,  we employed ResNet50 \cite{he2016deep} and TSM \cite{lin2019tsm} as RGB video backbone networks, while we used STGCN \cite{yu2018spatio} for skeleton data and DeepConvLSTM \cite{singh2020deep} for inertial data. 
For the RGB video data, ResNet50 was used to extract features from each of eight video frames of  $\textrm{height} \times \textrm{width} \times \textrm{channel} = 224 \times 224 \times 3$.  TSM was used to extract features from the eight video frames of $\mathrm{time} \times \mathrm{height} \times \mathrm{width} \times \mathrm{channel} = 8 \times 224 \times 224 \times 3$. The STGCN method extracted features from skeleton data of $\textrm{joint} \times \textrm{spatial position} = 20 \times 3$ using a window of size 15 over 8 time steps. For the inertial sensor data with $\textrm{Accelerometer} \times \textrm{Rotation} = 3 \times 3$, DeepConvLSTM was used with a window size of 100 over 8 time steps. The unified Transformer was configured with parameters, including a dimension of 512 and 8 attention heads.
Each layer was repeated 4 times for both the FTMT-Sim and FTMT-Seq modules.
During training, we employed stochastic gradient descent (SGD) as the optimizer for the entire network, utilizing a momentum of 0.9 and a learning rate of 0.0005. The training process for the FTMT-Sim and FTMT-Seq modules lasted for 16 hours and 300 epochs on the UTD-MHAD dataset. All training and testing were conducted on a single NVIDIA Geforce GTX 1080Ti GPU, with a batch size of 8.

\textbf{NTU RGB+D:} 
In the case of UTD-MHAD, we used TSM \cite{lin2019tsm} to extract the features from RGB video of size $\mathrm{time} \times \mathrm{height} \times \mathrm{width} \times \mathrm{channel} = 8 \times 224 \times 224 \times 3$. We also used STGCN to extract the features from skeleton data with dimension $\mathrm{joint} \times \mathrm{spatial}\,\,\mathrm{position} \times \mathrm{person} = 25 \times 3 \times 2$. We used a window of size 15 over 8 time steps. 
Both the FTMT-Sim and FTMT-Seq modules were configured with a dimension of 512 and 16 attention heads.
The Transformer encoder layer was repeated 4 times.
For optimization, we employed stochastic gradient descent (SGD) with a momentum of 0.9 and a learning rate of 0.003. The training duration was set to 480 hours and 70 epochs for the FTMT-Sim module, and 502 hours and 50 epochs for the FTMT-Seq module. Training is conducted with two NVIDIA Geforce GTX 1080Ti GPUs, with a batch size of 8.

\begin{table}
\begin{center}
\caption{Performance comparison on UTD-MHAD dataset: Top-1 Accuracy}
\label{tab1:UTD_MHAD}
\begin{adjustbox}{width=0.45\textwidth}
\begin{tabular}{c | c | c | c}
\Xhline{2.5\arrayrulewidth}
Method & Year & Modality &\begin{tabular}[c]{@{}c@{}}Accuracy\\(\%)\end{tabular} \\ \hline \hline
DCNN \cite{imran2016human} & 2016 & RGB, Depth & 91.20 \\
MCRL \cite{liu2018rgb} & 2018 & RGB, Depth & 93.02 \\
PoseMap \cite{liu2018recognizing} & 2018 & 3D Skeleton, Heatmap & 94.51 \\
HDM \cite{zhao2019bayesian} & 2019 & RGB, Skeleton & 92.8 \\ 
CC \cite{peng2019correlation} & 2019 & RGB, Inertial & 94.87 \\
Gimme$^{\cdot}$ Signals \cite{memmesheimer2020gimme} & 2020 & Skeleton, Inertial, Wifi & 93.33 \\ 
TPSMMs \cite{chen2020convnets} & 2020 & Skeleton & 93.26\\
LSTM-CNN \cite{zhu2020exploring}&2020 &Skeleton & 93.2 \\ 
HAMLET \cite{islam2020hamlet} & 2020 & RGB, Depth, Skeleton, Inertial & 95.12 \\
Fusion-GCN \cite{duhme2021fusion} & 2022 & RGB, Skeleton, Inertial & 94.42 \\
VSKD \cite{ni2022cross} & 2022 & RGB, Inertial & 96.97 \\
MuMu \cite{islam2022mumu} & 2022 & RGB, Depth, Skeleton, Inertial & 97.60 \\
MAVEN \cite{islam2022maven} & 2022 & RGB, Depth, Skeleton, Inertial & 97.81 \\
Multi-stream CNNs \cite{khezerlou2023multi} & 2023 & RGB, Depth, Skeleton, Inertial & 96.95 \\
\hline
 \textbf{UCFFormer-Sim-R(Ours)}& - & RGB, Skeleton, Inertial & \textbf{99.04} \\
 \textbf{UCFFormer-Sim-T(Ours)}& - & RGB, Skeleton, Inertial & \textbf{99.99} \\
 \textbf{UCFFormer-Seq-R(Ours)}& - & RGB, Skeleton, Inertial & \textbf{99.03} \\
 \textbf{UCFFormer-Seq-T(Ours)}& - & RGB, Skeleton, Inertial & \textbf{99.60} \\
\hline 
\Xhline{2.5\arrayrulewidth}
\end{tabular}   
\end{adjustbox}
\end{center}
\end{table}

\begin{table}
\begin{center}
\caption{Performance comparison on NTU RGB+D dataset: Cross-Subject (CS) and Cross-View (CV)}
\label{tab2:NTU_RGB+D}
\begin{adjustbox}{width=0.45\textwidth}
\begin{tabular}{c | c | c | c | c}
\Xhline{2.5\arrayrulewidth}
Method & Year & Modality & CS & CV \\ \hline \hline
PoseMap \cite{liu2018recognizing} & 2018 & RGB, Skeleton & 91.7 & 95.2 \\
TSMF \cite{bruce2021multimodal}  & 2020 & RGB, Skeleton & 92.5 & 97.4 \\
HAC \cite{davoodikakhki2020hierarchical} & 2020 & RGB, Skeleton & 95.66 & 98.79 \\
VPN \cite{das2020vpn} & 2020 & RGB, Skeleton & 95.5 & 98.0 \\
FUSION  \cite{de2020infrared} & 2020 & IR, Skeleton & 91.8 & 94.9 \\
InfoGCN \cite{chi2022infogcn}  & 2021 & Skeleton & 93.0 & 97.1 \\
PoseC3D \cite{duan2022revisiting} & 2021 & RGB, Skeleton & 97.0 & \textbf{99.6} \\
STAR-Transformer \cite{ahn2023star} & 2022 & RGB, Skeleton & 92.0 & 96.5 \\
\hline
\textbf{UCFFormer-Sim-T(Ours)} & - & RGB, Skeleton & \textbf{97.1} &  99.3 \\
\textbf{UCFFormer-Seq-T(Ours)} & - & RGB, Skeleton & 95.0 & 98.0 \\

\Xhline{2.5\arrayrulewidth}
\end{tabular}   
\end{adjustbox}
\end{center}
\end{table}

\begin{figure*}[tbh]
    \centering
    \captionsetup[subfloat]{labelfont=tiny,textfont=tiny}
    \subfloat[RGB]{ 
        \includegraphics[width=.13\textwidth]{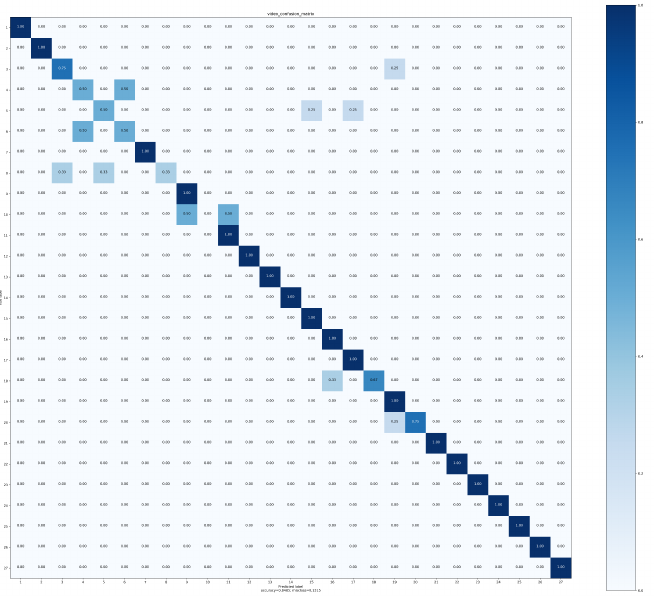}}
    \quad
    \subfloat[Inertial]{ 
        \includegraphics[width=.13\textwidth]{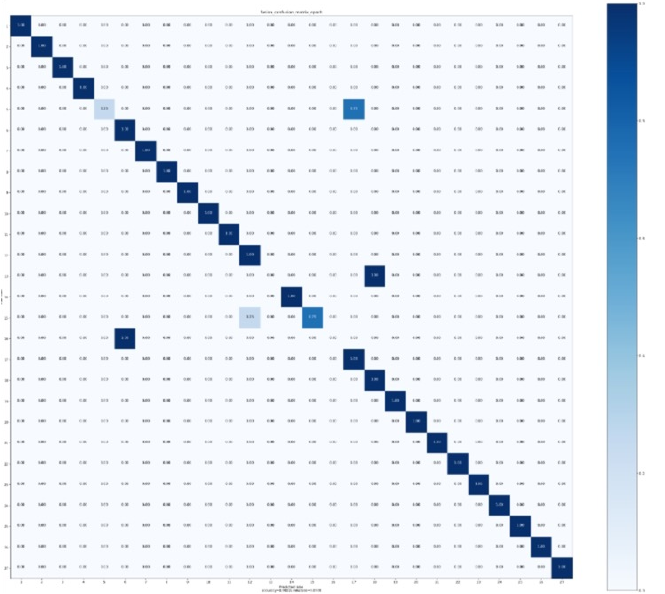}}
    \quad
    \subfloat[Skeleton]{ 
        \includegraphics[width=.13\textwidth]{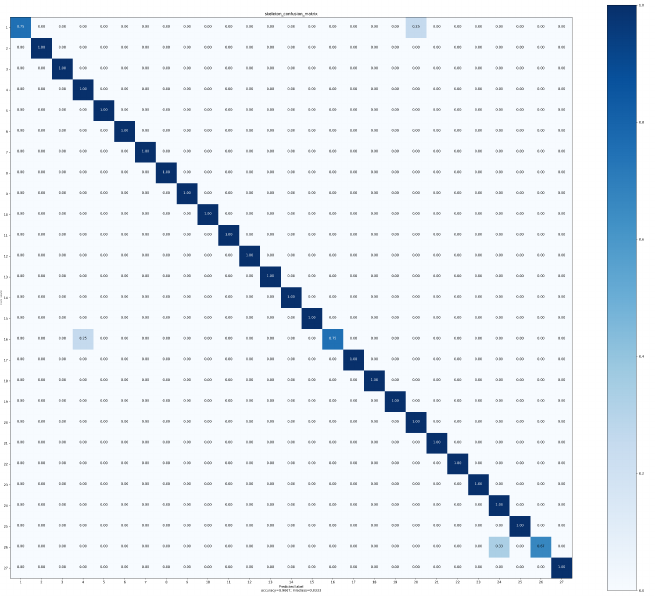}}
    \quad
    \subfloat[RGB+Inertial]{ 
        \includegraphics[width=.13\textwidth]{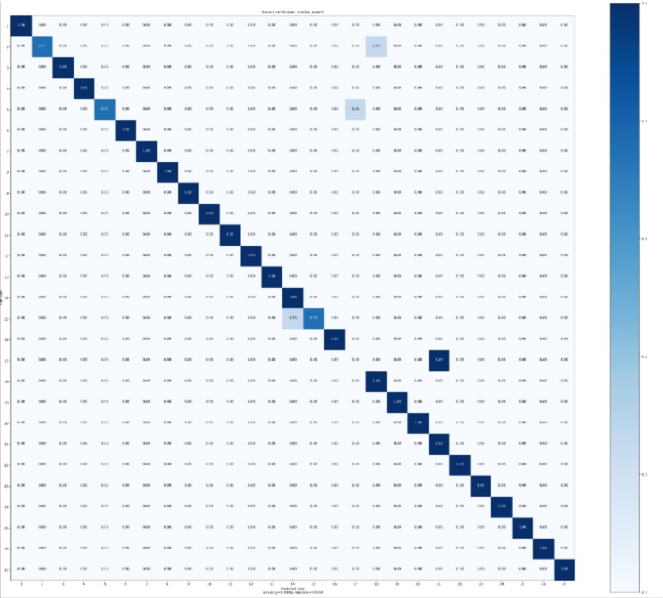}}
    \quad
    \subfloat[RGB+Skeleton]{ 
        \includegraphics[width=.13\textwidth]{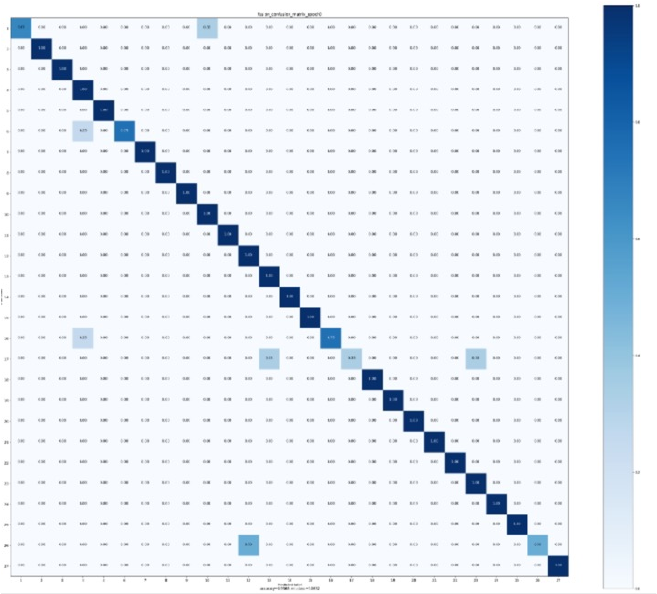}}
    \quad
    \subfloat[RGB+Inertial+Skeleton]{ 
        \includegraphics[width=.13\textwidth]{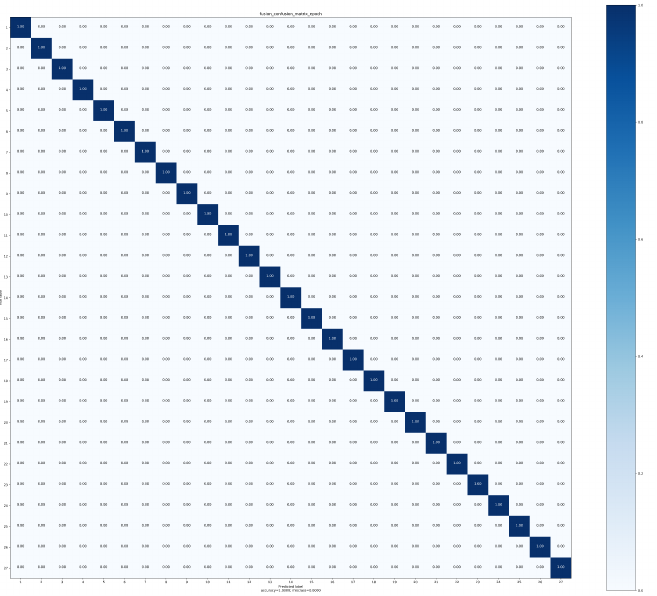}}
    \caption{\textbf{Confusion matrices for demonstrating the effect of multimodal fusion}. The evaluation is conducted using UCFFormer-Sim-R architecture on the UTD-MHAD dataset. 
    }
    \label{fig4:confusion_matrix}
\end{figure*}

\subsection{Performance Comparison}
The performance of UCFFormer was evaluated in comparison with conventional HAR methods. 
Table \ref{tab1:UTD_MHAD} presents the Top 1 accuracy results of the HAR models of interest evaluated on the UTD-MHAD dataset. 
We note that all UCFFormer variants achieve significant performance gains over the other HAR methods. In particular, UCFFormer-Sim-T set a new state-of-the-art performance with an impressive Top 1 accuracy of 99.99\%. It outperforms the latest MAVEN model \cite{islam2022maven} by a significant margin of 2.18\%.  These results show the effectiveness of UCFFormer in handling the multimodal data. We also observe that both FTMT-Sim and FTMT-Seq achieve comparable performance. 

\begin{table}
\begin{center}
\caption{Performance impact of multimodal fusion evaluated  on the UTD-MHAD Dataset}
\label{tab3:Modality_Test}
\begin{adjustbox}{width=0.45\textwidth}
\begin{tabular}{c||A|A|A||c}
\Xhline{2.5\arrayrulewidth}
 \multirow{2}{*}{Method} & \multicolumn{3}{c||}{Modalities} &\multirow{2}{*}{\begin{tabular}[c]{@{}c@{}}Accuracy\\(\%)\end{tabular}}\\\cline{2-4}
 & RGB & Inertial & Skeleton & \\
\hline\hline
\multirow{3}{*}{\begin{tabular}[c]{@{}c@{}}
Unimodal\end{tabular}} & \checkmark & & & 87.54\\
& & \checkmark & & 86.00   \\
& & & \checkmark &  94.72 \\ \hline
\multirow{3}{*}{\begin{tabular}[c]{@{}c@{}}
Multimodal\end{tabular}} & \checkmark & \checkmark & & 91.35 \\
& \checkmark &  & \checkmark & 96.15 \\
& \checkmark & \checkmark & \checkmark & 99.04 \\
\Xhline{2.5\arrayrulewidth}
\end{tabular}   
\end{adjustbox}
\end{center}
\end{table}

Table \ref{tab2:NTU_RGB+D} presents the performance of UCFFormer evaluated on the NTU RGB+D dataset \cite{shahroudy2016ntu}.
UCFFormer exhibits a performance that is on par with the current state-of-the-art method, PoseC3D \cite{duan2022revisiting}. Notably, UCFFormer offers substantial performance gains over other HAR models, underscoring its competitive performance in accurately recognizing human activities. It is also worth mentioning that FTMT-Sim achieves slightly better performance than FTMT-Seq on NTU RGB+D dataset. 

\subsection{Performance Behavior}
In this section, we present a comprehensive analysis of performance behavior of our UCFFormer.

\subsubsection{Effect of Multimodal Fusion}
Table \ref{tab3:Modality_Test} investigates performance gains achieved by multimodal fusion using UCFFormer-Sim-R on the UTD-MHAD dataset. 
While the original UCFFormer combined three modalities including RGB video, skeleton, and inertial data, Table \ref{tab3:Modality_Test} provides the performance achieved when only a single modality is used or only two modalities are combined. When a single modality was used, the embedding vectors were encoded solely through Temporal Attention, and then the flattened embedding vectors were used for classification. We observe that the performance of UCFFormer considerably decreases when employing a single modality or combining two modalities. When we use RGB, inertial, and skeleton data individually, the skeleton data yields the highest performance 94.72\%, which is 4.32\% lower than the accuracy of three modalities. When only two modalities are combined, the highest achieved accuracy is 96.15\%, exhibiting a 2.89\% reduction compared to the accuracy of the original design.
Figure \ref{fig4:confusion_matrix} presents the confusion matrices obtained with different combinations of modalities used for UCFFormer-Sim-R. It confirms that the inclusion of an additional modality leads to substantial improvements in classification accuracy.

\begin{figure}[!t]
    \centering
    \captionsetup[subfloat]{labelfont=footnotesize,textfont=footnotesize}
    \subfloat[The t-SNE plot with contrastive learning]{ 
        \includegraphics[width=.4\textwidth]{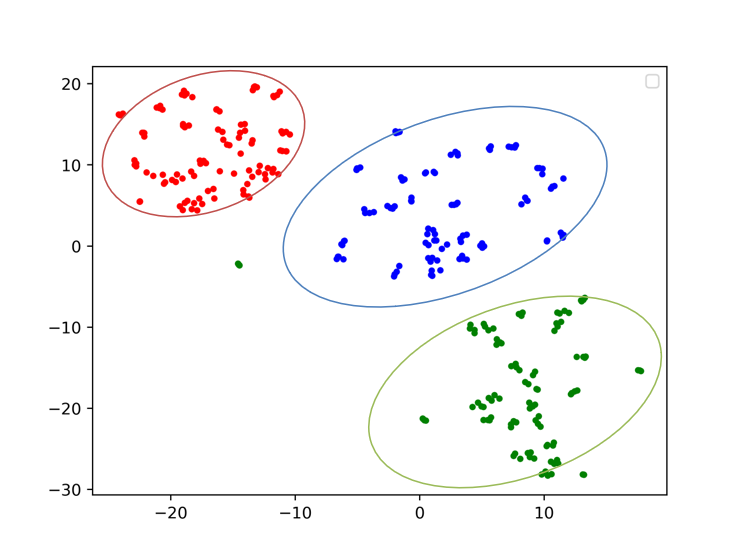}}
    \quad
    \subfloat[The t-SNE plot without constrative learning]{ 
        \includegraphics[width=.4\textwidth]{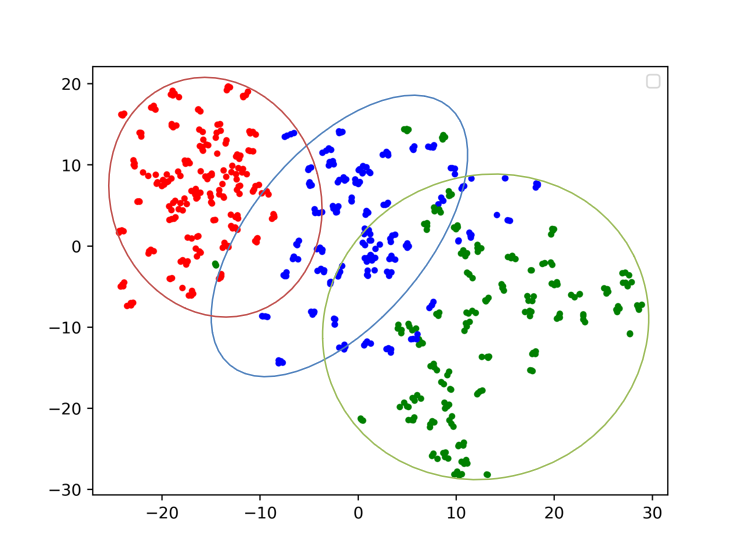}}
         \vspace{2mm}
    \caption{\textbf{t-SNE visualization of multimodal feature vectors.} We present the t-SNE plots of the multimodal feature vectors extracted from the RGB video (represented in red), skeleton (represented in blue), and inertial (represented in green) modalities. We compare the feature distributions with contrastive learning versus without contrastive learning.
    }
    \label{fig6:tsne}
\end{figure}

\begin{figure}[thb]
    \centering
    \captionsetup[subfloat]{labelfont=footnotesize,textfont=footnotesize}
    \subfloat[Performance versus \# of Transformer layers]{ 
        \includegraphics[width=.33\textwidth]{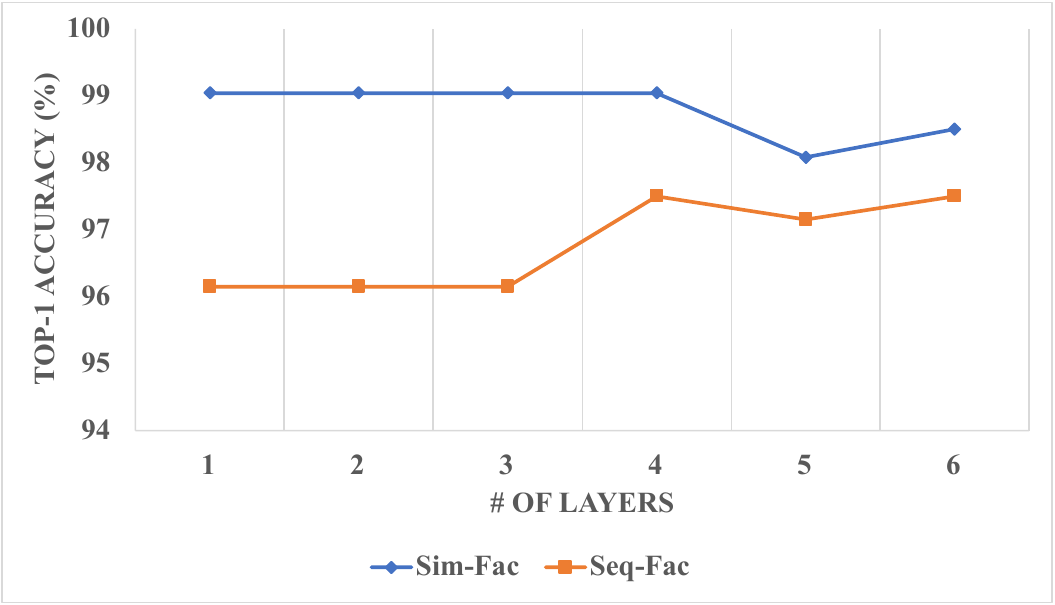}}
        \vspace{5mm}
     \subfloat[Performance versus \# of multi-attention heads]{ 
        \includegraphics[width=.33\textwidth]{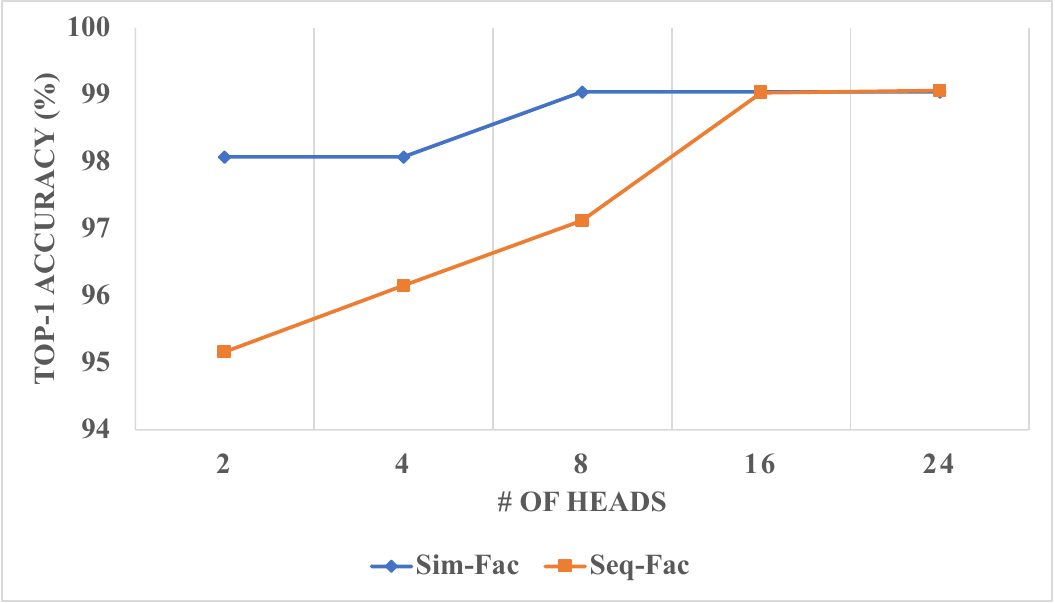}}
     \vspace{5mm}
    \subfloat[Performance versus embedding size]{ 
        \includegraphics[width=.33\textwidth]{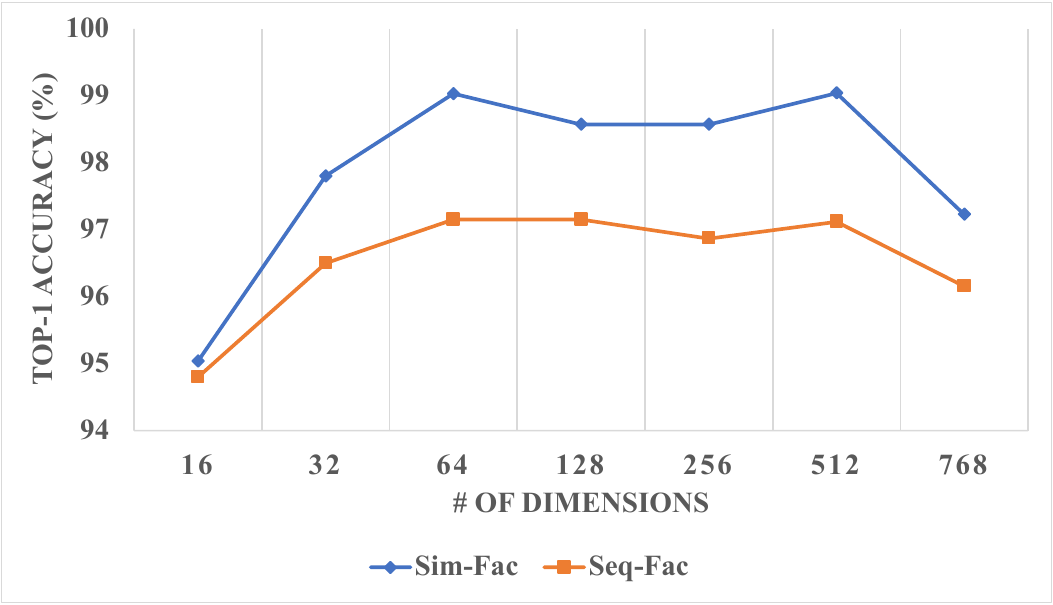}}
        \vspace{2mm}
    \caption{Performance of UCFFormer for different hyperparameter values. }
    \label{fig5:transformer_parameter}
\end{figure}

\begin{table}[tbh]
\begin{center}
\caption{Ablation study on UTD-MHAD dataset (Top-1 Accuracy)}
\label{tab4:Module_Test}
\begin{adjustbox}{width=0.35\textwidth}
\begin{tabular}{A|A|A||c}
\Xhline{2.5\arrayrulewidth}
 \multicolumn{3}{c||}{Settings} &\multirow{2}{*}{\begin{tabular}[c]{@{}c@{}}Accuracy\\(\%)\end{tabular}}\\\cline{1-3}
 {\begin{tabular}[c]{@{}c@{}}FTMT-Sim\end{tabular}} & {\begin{tabular}[c]{@{}c@{}}FTMT-Fac\end{tabular}} & MCANet  & \\
\hline\hline
& & & 92.03 \\

\checkmark & & & 97.87\\ 
 & \checkmark & &  97.13 \\
 &  & \checkmark & 96.15 \\
\hline
 \checkmark & & \checkmark & 99.04 \\
 & \checkmark & \checkmark & 99.03 \\
\Xhline{2.5\arrayrulewidth}
\end{tabular}   
\end{adjustbox}
\end{center}
\end{table}

\subsubsection{Effect of Contrastive Learning}

Figure \ref{fig6:tsne} displays a t-SNE analysis that represents features in a reduced-dimensional space. We compared the feature distributions when using contrastive learning and when not using it.
We observed that with the application of contrastive learning, the distribution gap in features across different modalities is notably reduced as intended.

\subsubsection{Performance versus Hyperparameters}
In Figure \ref{fig5:transformer_parameter}, we evaluate the performance of the UCFFormer when applying different hyperparameter values including  the number of Transformer layers, embedding size, and attention heads.   The default values of the number of layers, the embedding size, and the number of attention heads are set to 4, 512, and 8, respectively.

Figure \ref{fig5:transformer_parameter} (a) presents the performance versus the number of Transformer layers for both FTMT-Sim and FTMT-Seq. Our observations reveal that the peak classification accuracy is attained with 4 layers, and increasing the depth beyond this point adversely impacts performance. 
Figure \ref{fig5:transformer_parameter} (b) provides the performance versus the number of multi-heads. Performance improvement diminishes as the number of heads is above 8. Figure \ref{fig5:transformer_parameter} (c) provides the performance of UCFFormer versus embedding size. The performance dramatically improves as embedding size increases from 16 up to 64 but plateaus beyond 64, reaching its peak at 512. For all cases, FTMT-Sim consistently outperforms FTMT-Seq. 

\subsubsection{Ablation Studies} 
Table \ref{tab4:Module_Test} presents an ablation study presenting the contributions made by each component of UCFFormer to the overall performance.  We employed a ResNet-50 backbone and evaluated the performance using the Top-1 accuracy metric on the UTD-MHAD dataset. The baseline method neither includes FTMT nor MCANet; it directly inputs the embedding vectors into the Flatten + MLP module.  Table \ref{tab4:Module_Test} demonstrates that the baseline attains an accuracy of merely 92.03\%, which falls short by 7\% in comparison to UCFFormer-Sim-R. Upon integrating FTMT-Sim with the baseline, there is a notable accuracy enhancement of 5.84\%. The inclusion of FTMT-Seq results in a performance improvement of 5.1\%. This demonstrates that the unified Transformer enhances the action semantics through joint time-modality self-attention. Adding MCANet further improves the accuracy by 1.17\% for FTMT-Sim and by 1.9\% for FTMT-Seq. Note that in the absence of FTMT, the utilization of MCANet alone yields a notable 4.12\% improvement in accuracy compared to the baseline, which demonstrates the efficacy of MCANet.

\begin{table}[t]
    \begin{center}
    \caption{Comparison of computational complexity between factorized Self-Attention versus full self-attention on the UTD-MHAD dataset}
    \label{tab5:Comp_Attention}
    \resizebox{\columnwidth}{!}{%
        \begin{tabular}{ c || c | c | A | A | A | A }
        \Xhline{2.5\arrayrulewidth}
 \multirow{2}{*}{Module} & \multirow{2}{*}{\begin{tabular}{c} \# of \\ layers \end{tabular}} & \multirow{2}{*}{Accuracy (\%)} & \multicolumn{2}{c|}{Whole Network} & \multicolumn{2}{c}{ Transformer Encoder} \\\cline{4-7}
 & & & Params(M) & GFLOPs & Params(M)  & MFLOPs\\
        \hline \hline
        \multirow{4}{*}{\begin{tabular}[c]{@{}c@{}}Factorized \\Self-Attention \end{tabular}} & 2 & 99.04 & 33.72 & 66.96 & 7.69 & 122.8\\ \cline{2-7}
          & 3 & 99.04 & 36.87 & 67.00 & 10.84 & 173.16 \\ \cline{2-7}
          & 4 & 99.04 & 40.02 & 67.06 & 13.99 & 223.52 \\ \cline{2-7}
          & 5 & 98.07 & 43.17 & 67.06 & 17.14 & 273.88 \\
        \hline 
        \multirow{4}{*}{Self-Attention}& 2 & 97.15 & 43.17 & 67.1 & 17.15 & 273.94 \\ \cline{2-7}
          & 3 & 96.12 & 51.05 & 67.22 & 25.03 & 399.88 \\ \cline{2-7}
          & 4 & 95.04 & 58.93 & 67.36 & 32.91 & 525.82\\ \cline{2-7}
          & 5 & 87.75 & 66.81 & 67.48 & 40.79 & 705.74\\
        \Xhline{2.5\arrayrulewidth}
        \end{tabular}
        }
    \end{center}
\end{table}

Table \ref{tab5:Comp_Attention} compares the performance and computational complexity between the full-dimenional self-attention versus the factorized self-attention of FTMT-Sim.  
Table \ref{tab5:Comp_Attention} highlights the computational advantage of that Factorized Self-Attention over the conventional Self-Attention mechanism. For each added layer, Factorized Self-Attention adds approximately 50 million floating-point operations per second (MFLOPs), while the conventional Self-Attention sees an increase of 130 MFLOPs. 
Table \ref{tab5:Comp_Attention} also shows a substantial reduction in the number of parameters by adopting the Factorized Self-Attention. 
In spite of significant reduction in the memory usage and computation time, the factorized self-attention achieves slightly better accuracy than the full self-attention. 

\section{Conclusions}\label{sec5}
In this paper, we presented the UCFFormer, an innovative approach to feature-level fusion designed for HAR task. Employing two core components, FTMT  and MCANet, the UCFFormer established a solid framework for achieving effective multimodal fusion.
UCFFormer incrementally refines the embedding features extracted from each modality, leveraging both FTMT and MCANet.
FTMT captures the high-level inter-dependencies of embedding features spanning both time and modality domains. Utilizing the Factorized Time-Modality Self-attention mechanism, FTMT offers an efficient architecture for encoding multimodal features. MCANet then further refines these embedding features, employing contrastive loss to mitigate potential domain discrepancies that might arise among different modalities.
The performance of UCFFormer was evaluated on two widely used benchmarks. UCFFormer achieved the state-of-the-art performance, surpassing the latest HAR methods.
Ablation studies confirmed the effectiveness of the ideas applied to UCFFormer. 
In conclusion, the UCFFormer presents a robust and adaptable technique for merging various data types to enhance HAR performance. Beyond HAR, this research holds promise for other applications where a joint representation of diverse data types is essential for task execution.

\bibliographystyle{IEEEtran}
\bibliography{Reference}    

\begin{thebibliography}{10}
\providecommand{\url}[1]{#1}
\csname url@samestyle\endcsname
\providecommand{\newblock}{\relax}
\providecommand{\bibinfo}[2]{#2}
\providecommand{\BIBentrySTDinterwordspacing}{\spaceskip=0pt\relax}
\providecommand{\BIBentryALTinterwordstretchfactor}{4}
\providecommand{\BIBentryALTinterwordspacing}{\spaceskip=\fontdimen2\font plus
\BIBentryALTinterwordstretchfactor\fontdimen3\font minus
  \fontdimen4\font\relax}
\providecommand{\BIBforeignlanguage}[2]{{%
\expandafter\ifx\csname l@#1\endcsname\relax
\typeout{** WARNING: IEEEtran.bst: No hyphenation pattern has been}%
\typeout{** loaded for the language `#1'. Using the pattern for}%
\typeout{** the default language instead.}%
\else
\language=\csname l@#1\endcsname
\fi
#2}}
\providecommand{\BIBdecl}{\relax}
\BIBdecl

\bibitem{liu2022overview}
R.~Liu, A.~A. Ramli, H.~Zhang, E.~Henricson, and X.~Liu, ``An overview of human
  activity recognition using wearable sensors: Healthcare and artificial
  intelligence,'' in \emph{Internet of Things--ICIOT 2021: 6th International
  Conference, Held as Part of the Services Conference Federation, SCF 2021,
  Virtual Event, December 10--14, 2021, Proceedings}.\hskip 1em plus 0.5em
  minus 0.4em\relax Springer, 2022, pp. 1--14.

\bibitem{pathan2019machine}
N.~S. Pathan, M.~T.~F. Talukdar, M.~Quamruzzaman, and S.~A. Fattah, ``A machine
  learning based human activity recognition during physical exercise using
  wavelet packet transform of ppg and inertial sensors data,'' in \emph{2019
  4th International Conference on Electrical Information and Communication
  Technology (EICT)}.\hskip 1em plus 0.5em minus 0.4em\relax IEEE, 2019, pp.
  1--5.

\bibitem{saini2020human}
R.~Saini and V.~Maan, ``Human activity and gesture recognition: A review,'' in
  \emph{2020 International Conference on Emerging Trends in Communication,
  Control and Computing (ICONC3)}.\hskip 1em plus 0.5em minus 0.4em\relax IEEE,
  2020, pp. 1--2.

\bibitem{fan2022context}
L.~Fan, P.~Delir~Haghighi, Y.~Zhang, A.~R.~M. Forkan, and P.~P. Jayaraman,
  ``Context-aware human activity recognition (ca-har) using smartphone built-in
  sensors,'' in \emph{Advances in Mobile Computing and Multimedia Intelligence:
  20th International Conference, MoMM 2022, Virtual Event, November 28--30,
  2022, Proceedings}.\hskip 1em plus 0.5em minus 0.4em\relax Springer, 2022,
  pp. 108--121.

\bibitem{kumari2017increasing}
P.~Kumari, L.~Mathew, and P.~Syal, ``Increasing trend of wearables and
  multimodal interface for human activity monitoring: A review,''
  \emph{Biosensors and Bioelectronics}, vol.~90, pp. 298--307, 2017.

\bibitem{cippitelli2017human}
E.~Cippitelli, E.~Gambi, and S.~Spinsante, ``Human action recognition with
  rgb-d sensors,'' \emph{Motion Tracking and Gesture Recognition}, vol.~97,
  2017.

\bibitem{taylor2010convolutional}
G.~W. Taylor, R.~Fergus, Y.~LeCun, and C.~Bregler, ``Convolutional learning of
  spatio-temporal features,'' in \emph{European conference on computer
  vision}.\hskip 1em plus 0.5em minus 0.4em\relax Springer, 2010, pp. 140--153.

\bibitem{qiu2017learning}
Z.~Qiu, T.~Yao, and T.~Mei, ``Learning spatio-temporal representation with
  pseudo-3d residual networks,'' in \emph{proceedings of the IEEE International
  Conference on Computer Vision}, 2017, pp. 5533--5541.

\bibitem{carreira2017quo}
J.~Carreira and A.~Zisserman, ``Quo vadis, action recognition? a new model and
  the kinetics dataset,'' in \emph{proceedings of the IEEE Conference on
  Computer Vision and Pattern Recognition}, 2017, pp. 6299--6308.

\bibitem{feichtenhofer2019slowfast}
C.~Feichtenhofer, H.~Fan, J.~Malik, and K.~He, ``Slowfast networks for video
  recognition,'' in \emph{Proceedings of the IEEE/CVF international conference
  on computer vision}, 2019, pp. 6202--6211.

\bibitem{diete2019vision}
A.~Diete, T.~Sztyler, and H.~Stuckenschmidt, ``Vision and acceleration
  modalities: Partners for recognizing complex activities,'' in \emph{2019 IEEE
  International Conference on Pervasive Computing and Communications Workshops
  (PerCom Workshops)}.\hskip 1em plus 0.5em minus 0.4em\relax IEEE, 2019, pp.
  101--106.

\bibitem{chen2014home}
C.~Chen, K.~Liu, R.~Jafari, and N.~Kehtarnavaz, ``Home-based senior fitness
  test measurement system using collaborative inertial and depth sensors,'' in
  \emph{2014 36th Annual International Conference of the IEEE Engineering in
  Medicine and Biology Society}.\hskip 1em plus 0.5em minus 0.4em\relax IEEE,
  2014, pp. 4135--4138.

\bibitem{khandnor2017survey}
P.~Khandnor, N.~Kumar \emph{et~al.}, ``A survey of activity recognition process
  using inertial sensors and smartphone sensors,'' in \emph{2017 International
  Conference on Computing, Communication and Automation (ICCCA)}.\hskip 1em
  plus 0.5em minus 0.4em\relax IEEE, 2017, pp. 607--612.

\bibitem{zhao2018deep}
Y.~Zhao, R.~Yang, G.~Chevalier, X.~Xu, and Z.~Zhang, ``Deep residual bidir-lstm
  for human activity recognition using wearable sensors,'' \emph{Mathematical
  Problems in Engineering}, vol. 2018, pp. 1--13, 2018.

\bibitem{al2018hierarchical}
M.~Al-Naser, H.~Ohashi, S.~Ahmed, K.~Nakamura, T.~Akiyama, T.~Sato, P.~X.
  Nguyen, and A.~Dengel, ``Hierarchical model for zero-shot activity
  recognition using wearable sensors.'' in \emph{ICAART (2)}, 2018, pp.
  478--485.

\bibitem{gao2021danhar}
W.~Gao, L.~Zhang, Q.~Teng, J.~He, and H.~Wu, ``Danhar: Dual attention network
  for multimodal human activity recognition using wearable sensors,''
  \emph{Applied Soft Computing}, vol. 111, p. 107728, 2021.

\bibitem{dang2020sensor}
L.~M. Dang, K.~Min, H.~Wang, M.~J. Piran, C.~H. Lee, and H.~Moon,
  ``Sensor-based and vision-based human activity recognition: A comprehensive
  survey,'' \emph{Pattern Recognition}, vol. 108, p. 107561, 2020.

\bibitem{ravi2016deep}
D.~Ravi, C.~Wong, B.~Lo, and G.-Z. Yang, ``Deep learning for human activity
  recognition: A resource efficient implementation on low-power devices,'' in
  \emph{2016 IEEE 13th international conference on wearable and implantable
  body sensor networks (BSN)}.\hskip 1em plus 0.5em minus 0.4em\relax IEEE,
  2016, pp. 71--76.

\bibitem{atrey2010multimodal}
P.~K. Atrey, M.~A. Hossain, A.~El~Saddik, and M.~S. Kankanhalli, ``Multimodal
  fusion for multimedia analysis: a survey,'' \emph{Multimedia systems},
  vol.~16, pp. 345--379, 2010.

\bibitem{moencks2019adaptive}
M.~Moencks, V.~De~Silva, J.~Roche, and A.~Kondoz, ``Adaptive feature processing
  for robust human activity recognition on a novel multi-modal dataset,''
  \emph{arXiv preprint arXiv:1901.02858}, 2019.

\bibitem{shahroudy2017deep}
A.~Shahroudy, T.-T. Ng, Y.~Gong, and G.~Wang, ``Deep multimodal feature
  analysis for action recognition in rgb+ d videos,'' \emph{IEEE transactions
  on pattern analysis and machine intelligence}, vol.~40, no.~5, pp.
  1045--1058, 2017.

\bibitem{islam2020hamlet}
M.~M. Islam and T.~Iqbal, ``Hamlet: A hierarchical multimodal attention-based
  human activity recognition algorithm,'' in \emph{2020 IEEE/RSJ International
  Conference on Intelligent Robots and Systems (IROS)}.\hskip 1em plus 0.5em
  minus 0.4em\relax IEEE, 2020, pp. 10\,285--10\,292.

\bibitem{islam2022mumu}
M.~M. Islam and T.~Iqba, ``Mumu: Cooperative multitask learning-based guided
  multimodal fusion,'' in \emph{Proceedings of the AAAI Conference on
  Artificial Intelligence}, vol.~36, no.~1, 2022, pp. 1043--1051.

\bibitem{ni2022cross}
J.~Ni, R.~Sarbajna, Y.~Liu, A.~H. Ngu, and Y.~Yan, ``Cross-modal knowledge
  distillation for vision-to-sensor action recognition,'' in \emph{ICASSP
  2022-2022 IEEE International Conference on Acoustics, Speech and Signal
  Processing (ICASSP)}.\hskip 1em plus 0.5em minus 0.4em\relax IEEE, 2022, pp.
  4448--4452.

\bibitem{islam2022maven}
M.~M. Islam, M.~S. Yasar, and T.~Iqbal, ``Maven: A memory augmented recurrent
  approach for multimodal fusion,'' \emph{IEEE Transactions on Multimedia}, pp.
  1--1, 2022.

\bibitem{zhou2020unified}
L.~Zhou, H.~Palangi, L.~Zhang, H.~Hu, J.~Corso, and J.~Gao, ``{Unified
  vision-language pre-training for image captioning and VQA},'' in
  \emph{Proceedings of the AAAI conference on artificial intelligence},
  vol.~34, no.~07, 2020, pp. 13\,041--13\,049.

\bibitem{chen2015utd}
C.~Chen, R.~Jafari, and N.~Kehtarnavaz, ``Utd-mhad: A multimodal dataset for
  human action recognition utilizing a depth camera and a wearable inertial
  sensor,'' in \emph{2015 IEEE International conference on image processing
  (ICIP)}.\hskip 1em plus 0.5em minus 0.4em\relax IEEE, 2015, pp. 168--172.

\bibitem{shahroudy2016ntu}
A.~Shahroudy, J.~Liu, T.-T. Ng, and G.~Wang, ``Ntu rgb+ d: A large scale
  dataset for 3d human activity analysis,'' in \emph{Proceedings of the IEEE
  conference on computer vision and pattern recognition}, 2016, pp. 1010--1019.

\bibitem{dawar2018action}
N.~{Dawar} and N.~Kehtarnavaz, ``Action detection and recognition in continuous
  action streams by deep learning-based sensing fusion,'' \emph{IEEE Sensors
  Journal}, vol.~18, no.~23, pp. 9660--9668, 2018.

\bibitem{dawar2018data}
N.~Dawar, S.~Ostadabbas, and N.~Kehtarnavaz, ``Data augmentation in deep
  learning-based fusion of depth and inertial sensing for action recognition,''
  \emph{IEEE Sensors Letters}, vol.~3, no.~1, pp. 1--4, 2018.

\bibitem{imran2020evaluating}
J.~Imran and B.~Raman, ``Evaluating fusion of rgb-d and inertial sensors for
  multimodal human action recognition,'' \emph{Journal of Ambient Intelligence
  and Humanized Computing}, vol.~11, no.~1, pp. 189--208, 2020.

\bibitem{zou2019wifi}
H.~Zou, J.~Yang, H.~P. Das, H.~Liu, Y.~Zhou, and C.~J. Spanos, ``Wifi and
  vision multimodal learning for accurate and robust device-free human activity
  recognition,'' in \emph{2019 IEEE/CVF Conference on Computer Vision and
  Pattern Recognition Workshops (CVPRW)}.\hskip 1em plus 0.5em minus
  0.4em\relax IEEE, 2019, pp. 426--433.

\bibitem{tran2015learning}
D.~Tran, L.~Bourdev, R.~Fergus, L.~Torresani, and M.~Paluri, ``Learning
  spatiotemporal features with 3d convolutional networks,'' in
  \emph{Proceedings of the IEEE international conference on computer vision},
  2015, pp. 4489--4497.

\bibitem{long2018multimodal}
X.~Long, C.~Gan, G.~Melo, X.~Liu, Y.~Li, F.~Li, and S.~Wen, ``Multimodal
  keyless attention fusion for video classification,'' in \emph{Proceedings of
  the AAAI Conference on Artificial Intelligence}, vol.~32, no.~1, 2018.

\bibitem{zhou2018temporal}
B.~Zhou, A.~Andonian, A.~Oliva, and A.~Torralba, ``Temporal relational
  reasoning in videos,'' in \emph{Proceedings of the European conference on
  computer vision (ECCV)}, 2018, pp. 803--818.

\bibitem{bromley1993signature}
J.~Bromley, I.~Guyon, Y.~LeCun, E.~S{\"a}ckinger, and R.~Shah, ``Signature
  verification using a" siamese" time delay neural network,'' \emph{Advances in
  neural information processing systems}, vol.~6, 1993.

\bibitem{chopra2005learning}
S.~Chopra, R.~Hadsell, and Y.~LeCun, ``Learning a similarity metric
  discriminatively, with application to face verification,'' in \emph{2005 IEEE
  Computer Society Conference on Computer Vision and Pattern Recognition
  (CVPR'05)}, vol.~1.\hskip 1em plus 0.5em minus 0.4em\relax IEEE, 2005, pp.
  539--546.

\bibitem{hadsell2006dimensionality}
R.~Hadsell, S.~Chopra, and Y.~LeCun, ``Dimensionality reduction by learning an
  invariant mapping,'' in \emph{2006 IEEE Computer Society Conference on
  Computer Vision and Pattern Recognition (CVPR'06)}, vol.~2.\hskip 1em plus
  0.5em minus 0.4em\relax IEEE, 2006, pp. 1735--1742.

\bibitem{weinberger2009distance}
K.~Q. Weinberger and L.~K. Saul, ``Distance metric learning for large margin
  nearest neighbor classification.'' \emph{Journal of machine learning
  research}, vol.~10, no.~2, 2009.

\bibitem{collobert2008unified}
R.~Collobert and J.~Weston, ``A unified architecture for natural language
  processing: Deep neural networks with multitask learning,'' in
  \emph{Proceedings of the 25th international conference on Machine learning},
  2008, pp. 160--167.

\bibitem{chechik2010large}
G.~Chechik, V.~Sharma, U.~Shalit, and S.~Bengio, ``Large scale online learning
  of image similarity through ranking.'' \emph{Journal of Machine Learning
  Research}, vol.~11, no.~3, 2010.

\bibitem{he2020momentum}
K.~He, H.~Fan, Y.~Wu, S.~Xie, and R.~Girshick, ``Momentum contrast for
  unsupervised visual representation learning,'' in \emph{Proceedings of the
  IEEE/CVF conference on computer vision and pattern recognition}, 2020, pp.
  9729--9738.

\bibitem{chen2020improved}
X.~Chen, H.~Fan, R.~Girshick, and K.~He, ``Improved baselines with momentum
  contrastive learning,'' \emph{arXiv preprint arXiv:2003.04297}, 2020.

\bibitem{chen2020simple}
T.~Chen, S.~Kornblith, M.~Norouzi, and G.~Hinton, ``A simple framework for
  contrastive learning of visual representations,'' in \emph{International
  conference on machine learning}.\hskip 1em plus 0.5em minus 0.4em\relax PMLR,
  2020, pp. 1597--1607.

\bibitem{chen2020big}
T.~Chen, S.~Kornblith, K.~Swersky, M.~Norouzi, and G.~E. Hinton, ``Big
  self-supervised models are strong semi-supervised learners,'' \emph{Advances
  in neural information processing systems}, vol.~33, pp. 22\,243--22\,255,
  2020.

\bibitem{grill2020bootstrap}
J.-B. Grill, F.~Strub, F.~Altch{\'e}, C.~Tallec, P.~Richemond, E.~Buchatskaya,
  C.~Doersch, B.~Avila~Pires, Z.~Guo, M.~Gheshlaghi~Azar \emph{et~al.},
  ``Bootstrap your own latent-a new approach to self-supervised learning,''
  \emph{Advances in neural information processing systems}, vol.~33, pp.
  21\,271--21\,284, 2020.

\bibitem{kenton2019bert}
J.~D. M.-W.~C. Kenton and L.~K. Toutanova, ``Bert: Pre-training of deep
  bidirectional transformers for language understanding,'' in \emph{Proceedings
  of NAACL-HLT}, 2019, pp. 4171--4186.

\bibitem{brown2020language}
T.~Brown, B.~Mann, N.~Ryder, M.~Subbiah, J.~D. Kaplan, P.~Dhariwal,
  A.~Neelakantan, P.~Shyam, G.~Sastry, A.~Askell \emph{et~al.}, ``Language
  models are few-shot learners,'' \emph{Advances in neural information
  processing systems}, vol.~33, pp. 1877--1901, 2020.

\bibitem{dosovitskiy2020image}
A.~Dosovitskiy, L.~Beyer, A.~Kolesnikov, D.~Weissenborn, X.~Zhai,
  T.~Unterthiner, M.~Dehghani, M.~Minderer, G.~Heigold, S.~Gelly \emph{et~al.},
  ``An image is worth 16x16 words: Transformers for image recognition at
  scale,'' \emph{arXiv preprint arXiv:2010.11929}, 2020.

\bibitem{hu2021unit}
R.~Hu and A.~Singh, ``Unit: Multimodal multitask learning with a unified
  transformer,'' in \emph{Proceedings of the IEEE/CVF International Conference
  on Computer Vision}, 2021, pp. 1439--1449.

\bibitem{wang2021ufo}
J.~Wang, X.~Hu, Z.~Gan, Z.~Yang, X.~Dai, Z.~Liu, Y.~Lu, and L.~Wang, ``Ufo: A
  unified transformer for vision-language representation learning,''
  \emph{arXiv preprint arXiv:2111.10023}, 2021.

\bibitem{li2021uniformer}
K.~Li, Y.~Wang, G.~Peng, G.~Song, Y.~Liu, H.~Li, and Y.~Qiao, ``Uniformer:
  Unified transformer for efficient spatial-temporal representation learning,''
  in \emph{International Conference on Learning Representations}, 2021.

\bibitem{ma2022unitranser}
Z.~Ma, J.~Li, G.~Li, and Y.~Cheng, ``Unitranser: A unified transformer semantic
  representation framework for multimodal task-oriented dialog system,'' in
  \emph{Proceedings of the 60th Annual Meeting of the Association for
  Computational Linguistics (Volume 1: Long Papers)}, 2022, pp. 103--114.

\bibitem{vaswani2017attention}
A.~Vaswani, N.~Shazeer, N.~Parmar, J.~Uszkoreit, L.~Jones, A.~N. Gomez,
  {\L}.~Kaiser, and I.~Polosukhin, ``Attention is all you need,''
  \emph{Advances in neural information processing systems}, vol.~30, 2017.

\bibitem{lin2019tsm}
J.~Lin, C.~Gan, and S.~Han, ``Tsm: Temporal shift module for efficient video
  understanding,'' in \emph{Proceedings of the IEEE/CVF international
  conference on computer vision}, 2019, pp. 7083--7093.

\bibitem{imran2016human}
J.~Imran and P.~Kumar, ``Human action recognition using rgb-d sensor and deep
  convolutional neural networks,'' in \emph{2016 international conference on
  advances in computing, communications and informatics (ICACCI)}.\hskip 1em
  plus 0.5em minus 0.4em\relax IEEE, 2016, pp. 144--148.

\bibitem{liu2018rgb}
T.~Liu, J.~Kong, and M.~Jiang, ``Rgb-d action recognition using multimodal
  correlative representation learning model,'' \emph{IEEE Sensors Journal},
  vol.~19, no.~5, pp. 1862--1872, 2018.

\bibitem{yu2018spatio}
B.~Yu, H.~Yin, and Z.~Zhu, ``Spatio-temporal graph convolutional networks: a
  deep learning framework for traffic forecasting,'' in \emph{Proceedings of
  the 27th International Joint Conference on Artificial Intelligence}, 2018,
  pp. 3634--3640.

\bibitem{singh2020deep}
S.~P. Singh, M.~K. Sharma, A.~Lay-Ekuakille, D.~Gangwar, and S.~Gupta, ``Deep
  convlstm with self-attention for human activity decoding using wearable
  sensors,'' \emph{IEEE Sensors Journal}, vol.~21, no.~6, pp. 8575--8582, 2020.

\bibitem{he2016deep}
K.~He, X.~Zhang, S.~Ren, and J.~Sun, ``Deep residual learning for image
  recognition,'' in \emph{Proceedings of the IEEE conference on computer vision
  and pattern recognition}, 2016, pp. 770--778.

\bibitem{liu2018recognizing}
M.~Liu and J.~Yuan, ``Recognizing human actions as the evolution of pose
  estimation maps,'' in \emph{Proceedings of the IEEE Conference on Computer
  Vision and Pattern Recognition}, 2018, pp. 1159--1168.

\bibitem{zhao2019bayesian}
R.~Zhao, W.~Xu, H.~Su, and Q.~Ji, ``Bayesian hierarchical dynamic model for
  human action recognition,'' in \emph{Proceedings of the IEEE/CVF Conference
  on Computer Vision and Pattern Recognition}, 2019, pp. 7733--7742.

\bibitem{peng2019correlation}
B.~Peng, X.~Jin, J.~Liu, D.~Li, Y.~Wu, Y.~Liu, S.~Zhou, and Z.~Zhang,
  ``Correlation congruence for knowledge distillation,'' in \emph{Proceedings
  of the IEEE/CVF International Conference on Computer Vision}, 2019, pp.
  5007--5016.

\bibitem{memmesheimer2020gimme}
R.~Memmesheimer, N.~Theisen, and D.~Paulus, ``Gimme signals: Discriminative
  signal encoding for multimodal activity recognition,'' in \emph{2020 IEEE/RSJ
  International Conference on Intelligent Robots and Systems (IROS)}.\hskip 1em
  plus 0.5em minus 0.4em\relax IEEE, 2020, pp. 10\,394--10\,401.

\bibitem{chen2020convnets}
Y.~Chen, L.~Wang, C.~Li, Y.~Hou, and W.~Li, ``Convnets-based action recognition
  from skeleton motion maps,'' \emph{Multimedia Tools and Applications},
  vol.~79, pp. 1707--1725, 2020.

\bibitem{zhu2020exploring}
A.~Zhu, Q.~Wu, R.~Cui, T.~Wang, W.~Hang, G.~Hua, and H.~Snoussi, ``Exploring a
  rich spatial--temporal dependent relational model for skeleton-based action
  recognition by bidirectional lstm-cnn,'' \emph{Neurocomputing}, vol. 414, pp.
  90--100, 2020.

\bibitem{duhme2021fusion}
M.~Duhme, R.~Memmesheimer, and D.~Paulus, ``Fusion-gcn: Multimodal action
  recognition using graph convolutional networks,'' in \emph{DAGM German
  Conference on Pattern Recognition}.\hskip 1em plus 0.5em minus 0.4em\relax
  Springer, 2021, pp. 265--281.

\bibitem{khezerlou2023multi}
F.~Khezerlou, A.~Baradarani, M.~A. Balafar, and R.~G. Maev, ``Multi-stream cnns
  with orientation-magnitude response maps and weighted inception module for
  human action recognition,'' in \emph{2023 3rd International conference on
  Artificial Intelligence and Signal Processing (AISP)}.\hskip 1em plus 0.5em
  minus 0.4em\relax IEEE, 2023, pp. 1--5.

\bibitem{bruce2021multimodal}
X.~Bruce, Y.~Liu, and K.~C. Chan, ``Multimodal fusion via teacher-student
  network for indoor action recognition,'' in \emph{Proceedings of the AAAI
  Conference on Artificial Intelligence}, vol.~35, no.~4, 2021, pp. 3199--3207.

\bibitem{davoodikakhki2020hierarchical}
M.~Davoodikakhki and K.~Yin, ``Hierarchical action classification with network
  pruning,'' in \emph{Advances in Visual Computing: 15th International
  Symposium, ISVC 2020, San Diego, CA, USA, October 5--7, 2020, Proceedings,
  Part I 15}.\hskip 1em plus 0.5em minus 0.4em\relax Springer, 2020, pp.
  291--305.

\bibitem{das2020vpn}
S.~Das, S.~Sharma, R.~Dai, F.~Bremond, and M.~Thonnat, ``Vpn: Learning
  video-pose embedding for activities of daily living,'' in \emph{Computer
  Vision--ECCV 2020: 16th European Conference, Glasgow, UK, August 23--28,
  2020, Proceedings, Part IX 16}.\hskip 1em plus 0.5em minus 0.4em\relax
  Springer, 2020, pp. 72--90.

\bibitem{de2020infrared}
A.~M. De~Boissiere and R.~Noumeir, ``Infrared and 3d skeleton feature fusion
  for rgb-d action recognition,'' \emph{IEEE Access}, vol.~8, pp.
  168\,297--168\,308, 2020.

\bibitem{chi2022infogcn}
H.-g. Chi, M.~H. Ha, S.~Chi, S.~W. Lee, Q.~Huang, and K.~Ramani, ``Infogcn:
  Representation learning for human skeleton-based action recognition,'' in
  \emph{Proceedings of the IEEE/CVF Conference on Computer Vision and Pattern
  Recognition}, 2022, pp. 20\,186--20\,196.

\bibitem{duan2022revisiting}
H.~Duan, Y.~Zhao, K.~Chen, D.~Lin, and B.~Dai, ``Revisiting skeleton-based
  action recognition,'' in \emph{Proceedings of the IEEE/CVF Conference on
  Computer Vision and Pattern Recognition}, 2022, pp. 2969--2978.

\bibitem{ahn2023star}
D.~Ahn, S.~Kim, H.~Hong, and B.~C. Ko, ``Star-transformer: A spatio-temporal
  cross attention transformer for human action recognition,'' in
  \emph{Proceedings of the IEEE/CVF Winter Conference on Applications of
  Computer Vision}, 2023, pp. 3330--3339.

\end{thebibliography}

\end{document}